# Modified Hybrid A* Collision-Free Path-Planning for Automated Reverse Parking

Xincheng Cao, Haochong Chen, Bilin Aksun-Guvenc, Levent Guvenc

Automated Driving Lab, Ohio State University

## Abstract

Parking a vehicle in tight spaces is a challenging task to perform due to the scarcity of feasible paths that are also collision-free. This paper presents a strategy to tackle this kind of maneuver with a modified Hybrid-A* path-planning algorithm that combines the feasibility guarantee inherent in the standard Hybrid A* algorithm with the addition of static obstacle collision avoidance. A kinematic single-track model is derived to describe the low-speed motion of the vehicle, which is subsequently used as the motion model in the Hybrid A* path-planning algorithm to generate feasible motion primitive branches. The model states are also used to reconstruct the vehicle centerline, which, in conjunction with an inflated binary occupancy map, facilitates static obstacle collision avoidance functions. Simulation study and animation are set up to test the efficacy of the approach, and the proposed algorithm proves to consistently provide kinematically feasible trajectories that are also collision-free.

## Introduction

Vehicle automation has experienced remarkable growth over the past decade, driven largely by advances in sensing, computation, and the widespread deployment of Advanced Driver Assistance Systems (ADAS) [1–4]. These technologies have steadily shifted vehicles from passive safety systems toward increasingly autonomous decision-making capabilities. Among the core competencies required for higher levels of autonomy, the ability to generate, evaluate, and follow a planned path is fundamental [5]. Path planning enables a vehicle to navigate safely and efficiently through complex environments, while path tracking ensures that the planned trajectory is executed accurately despite disturbances or modeling uncertainties.

Within this broader context, parking maneuvers represent a particularly important yet challenging subset of autonomous driving tasks. Although parking is a routine activity for human drivers, it becomes significantly more complex for automated systems, especially in dense urban environments where available space is limited and maneuvering margins are small. Automated parking systems must not only generate feasible trajectories that respect vehicle kinematics but also ensure that these trajectories remain collision-free in environments cluttered with static obstacles such as parked vehicles, curbs, and infrastructure elements. As highlighted in [6], the ability to perform reliable and safe parking maneuvers is essential for improving user trust and enabling fully autonomous valet parking applications.

The difficulty of parking in tight spaces arises from the scarcity of feasible paths that simultaneously satisfy nonholonomic vehicle constraints and avoid collisions. This challenge has motivated extensive research into path-planning and control strategies tailored to low-speed, highly constrained environments. A wide range of methods has been explored in the literature. Comprehensive treatments of trajectory generation, path tracking, and the underlying vehicle dynamics can be found in [5], while robust mechatronic control strategies applicable to autonomous driving systems are detailed in [7]. Multi-objective PID controller design for automated path following was introduced in [8], offering a systematic approach to balancing competing performance criteria. Cooperative adaptive cruise control systems, which extend path-following concepts to multi-vehicle coordination, were developed in [9]. Additionally, foundational work on robust control theory linking parameter space and frequency domain methods was presented in [10], providing theoretical tools relevant to delay-tolerant and safety-critical autonomous architectures. As all applications of connected and autonomous driving have inherent time delays due to communication, perception, decision making and actuation, delay aware and tolerant control was studied in [11].

Collision avoidance remains one of the most critical aspects of autonomous navigation, particularly in constrained parking environments. Elastic band theory, originally proposed in [12], introduced a deformable-path approach that adapts trajectories in real time to avoid obstacles. This concept was later extended to pedestrian-rich environments using V2X communication in [13], demonstrating its applicability to low-speed autonomous shuttles. Further work in [14] examined collision avoidance strategies specifically for low-speed shuttle operations, highlighting the importance of perception fidelity and real-time responsiveness. Complementary research on vehicle stability, such as active steering systems [15] and yaw stability controllers [16], provides additional safety layers that can support precise maneuvering in narrow spaces.

Accurate perception and localization are equally essential for successful autonomous parking. Low-speed autonomous shuttle deployments described in [17] demonstrated the effectiveness of sensor fusion techniques for reliable state estimation. Similarly, the integration of camera and GPS data for lane keeping, as shown in [18], underscores the importance of combining multiple sensing modalities to achieve robust localization in environments where GPS alone may be insufficient. Broader smart mobility initiatives, such as the SmartShuttle project [19], situate autonomous parking within the larger ecosystem of connected and intelligent transportation systems, emphasizing scalability and integration with smart city infrastructure.

Before deploying autonomous parking algorithms in real vehicles, extensive simulation and testing are required to ensure safety and reliability. Scenario-based pre-deployment testing in simulated environments was highlighted in [20], demonstrating how virtual testing can accelerate development while reducing risk. Hardware-in-the-loop simulation frameworks for connected and autonomous vehicles, developed in [21], further bridge the gap between software validation and real-world implementation. Beyond ground vehicles, coordinated operation between UAVs and connected autonomous vehicles has been explored in [22], illustrating the growing interest in multi-agent systems for enhanced situational awareness and safety. These developments are complemented by broader discussions of connected and autonomous vehicles within the context of IoT and data analytics [2], which highlight the increasing role of data-driven intelligence in modern mobility systems.

The present study focuses on the development of a modified Hybrid A* path-planning algorithm specifically designed for automated reverse parking in tight spaces. While Hybrid A* is widely recognized for its ability to generate kinematically feasible trajectories, traditional implementations do not explicitly incorporate collision-checking mechanisms that account for the full vehicle geometry. This work addresses that gap by integrating a kinematic single-track model with an inflated occupancy map–based collision detection framework. The resulting algorithm ensures that each motion primitive generated during the search process is both kinematically feasible and guaranteed to be collision-free. Through simulation studies, we demonstrate that the proposed method consistently produces safe and feasible reverse parking trajectories even in highly constrained environments.

Building on this body of work, the present study introduces a modified Hybrid-A* path-planning algorithm integrated with a kinematic single-track model and static obstacle avoidance functions. By combining feasibility guarantees with collision-free trajectory generation, and validating through simulation and visualization in Simulink and Unreal Engine, this research advances autonomous valet parking capabilities.

Various general-purpose path-planning strategies exist in the literature. Examples include elastic band approach discussed in [23], potential field method illustrated in [24] and A* algorithm detailed in [25]. Further modifications of these classic methods have also been attempted. For example, the D* algorithm described in [26] and the Hybrid A* algorithm explained in [27] are both extensions of the original A* method. Meanwhile, [5] illustrates some additional approaches in describing a path using Bezier curves, clothoids, and polynomial splines. Some comparison studies also exist, such as [28] that compares the performance of a flatness-based planner, a polynomial trajectory planner and a symmetric polynomial trajectory planner.

Path-planning approaches tailored to the scenario of autonomous parking have also been explored in the existing literature. For example, [29] provides a comprehensive overview of autonomous parking system that summarizes two main families of path-planning approaches: optimization and control-based methods as well as search and sampling-based methods. Examples of optimization-based approaches include [30] that aims to achieve collision avoidance functionality of a coarse initial guess trajectory by introducing an optimization formulation, as well as [31] that recursively uses static optimization formulation on discretized path segments to generate a feasible parking trajectory. Examples of search and sampling-based approaches include [32] that proposes a reverse searching method based on the Hybrid A* algorithm to improve planning efficiency in cluttered parking environments, as well as [33] that features a modified RRT planner specifically tuned for narrow parking spaces. Apart from the above-mentioned two families of approaches, some additional planning methods also exist. For instance, [34] presents a geometric planning procedure where an initial path is constructed using circular arc segments to negotiate the narrow parking space in one or multiple maneuvers and then transform them into a smooth continuous-curvature path using clothoids. Similarly, [35] uses minimum radius circular arcs to construct the path to tackle parallel parking tasks with backward and forward (BF) maneuvers. While the above-referenced methods work well, they do not explicitly take into account collision avoidance with already parked vehicles next to tight parking spaces. This paper contributes to the existing literature by extending the hybrid A* method by checking for possible collisions as each step of the search and determining a collision free path during the reverse parking maneuver. While the derivation and results are for reverse parking, they can easily be modified and applied to head on parking.

The organization of the remainder of this paper is as follows. A kinematic single-track vehicle model is derived first. This is followed by the explanation of the modified Hybrid A* algorithm, including design details of cost function, motion primitive generation, collision check and priority queue contents. A simulation case study is then added to demonstrate the effectiveness of the proposed planning algorithm. The paper ends with conclusions.

## Kinematic Vehicle Model

Due to the low-speed nature of parking maneuvers, kinematic vehicle model is a suitable choice to be used as the motion model for the Hybrid A* algorithm. This section, hence, presents the derivation of a kinematic single-track model with front wheel steering. The model schematic is displayed in Figure 1, and the parameters of this model are listed in Table 1. If we assign vehicle rear axle center as the position reference point, we can apply the transport formula and derive the position equations of the front axle center as displayed in Equation 1 and 2.

$$X_F = X_R + L \cdot \cos(\psi) \quad (1)$$

$$Y_F = Y_R + L \cdot \sin(\psi) \quad (2)$$

Applying time derivatives to the axle positions yield velocity equations as displayed in Equation 3 and 4.

$$\vec{V_R} = \begin{pmatrix} \dot{X}_R \\ \dot{Y}_R \end{pmatrix} \quad (3)$$

$$\vec{V_F} = \begin{pmatrix} \dot{X}_F \\ \dot{Y}_F \end{pmatrix} = \begin{pmatrix} \dot{X}_R - L \cdot \dot{\psi} \cdot \sin(\psi) \\ \dot{Y}_R + L \cdot \dot{\psi} \cdot \cos(\psi) \end{pmatrix} \quad (4)$$

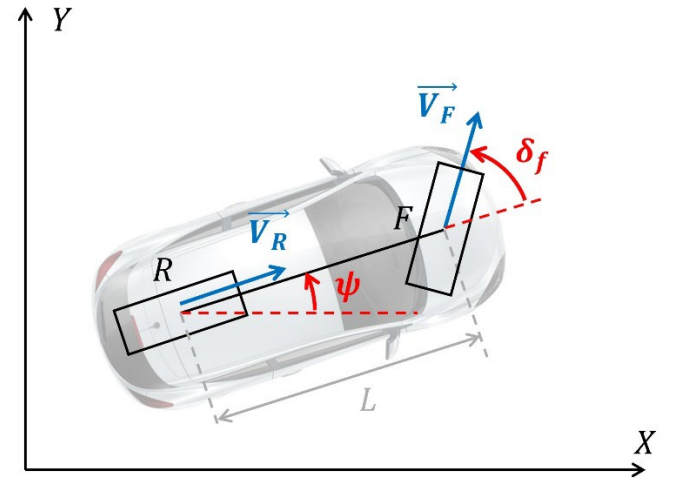

Figure 1. Kinematic single-track vehicle model

Table 1. Parameters of kinematic single-track vehicle model

| Model Parameter | Explanation |
| --- | --- |
| $L$ | Wheelbase of the vehicle |
| $\delta_f$ | Vehicle front wheel steer angle |
| $\psi$ | Vehicle yaw angle |
| $\overrightarrow{V_F}$ | Vehicle front axle velocity |
| $\overrightarrow{V_R}$ | Vehicle rear axle velocity |

It should be remarked that in Figure 1, the velocities of the vehicle front and rear wheels are aligned with their respective orientation. This is due to the tires with no side slip assumption since tire deformation at low speed is negligible and the direction of tire travel is assumed to follow the tire heading. This assumption induces kinematic constraints, and they can be represented as shown in Equation 5 and 6.

$$\tan(\psi) = \frac{V_R \cdot \sin(\psi)}{V_R \cdot \cos(\psi)} = \frac{\dot{Y}_R}{\dot{X}_R} \tag{5}$$

$$\tan(\psi + \delta_f) = \frac{\dot{Y}_F}{\dot{X}_F} \tag{6}$$

Note that the term $V_R$ is the magnitude of the vehicle rear axle center velocity $\overrightarrow{V_R}$ and can either be positive or negative depending on the vehicle direction of travel. Combining the velocity equations and the kinematic constraint equations and simplifying further, one can obtain the kinematic single-track vehicle model as shown in Equations 7 to 9.

$$\dot{X}_R = V_R \cdot \cos(\psi) \tag{7}$$

$$\dot{Y}_R = V_R \cdot \sin(\psi) \tag{8}$$

$$\dot{\psi} = \frac{V_R}{L} \tan(\delta_f) \tag{9}$$

In this model, the inputs are the vehicle front axle steer angle $\delta_f$ and vehicle rear axle center speed $V_R$.

## Modified Hybrid A* Path-Planning Algorithm

### Algorithm Overview

A typical vehicle reverse parking scenario is illustrated in Figure 2, where the path-planning goal is to position the vehicle into the desired parking space while avoiding collisions with any static obstacles that are usually located in neighboring parking spaces. It is worth noting that while collisions with any dynamic obstacles must also be avoided, the trivial solution of stopping the vehicle and waiting for them to pass is the safest and most common. As a result, dynamic obstacle avoidance is not included as one of the planning goals. The satisfactory completion of this path-planning operation requires simultaneous fulfillment of the following conditions: 1) the generated path must be kinematically feasible for the vehicle in reverse motion; 2) the generated path must be collision-free. Given these requirements, the Hybrid A* algorithm serves as a good starting point as it takes system kinematics into account by default and can be expanded to include static collision avoidance features. This sub-section hence provides a general overview of a modified Hybrid A* algorithm while later sub-sections explain further design details.

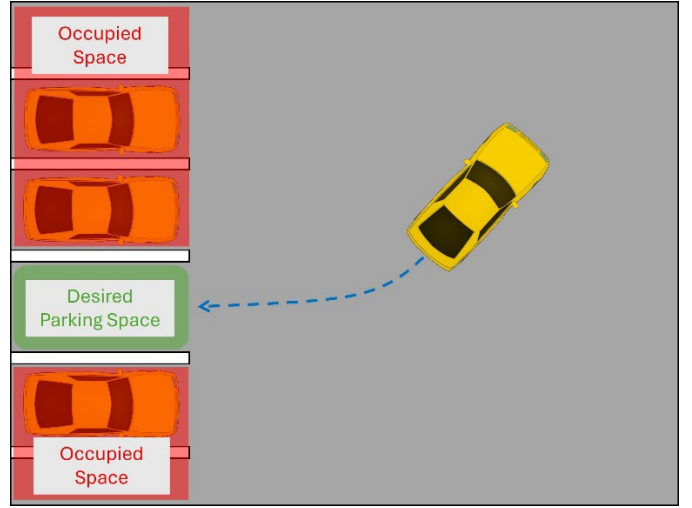

Figure 2. Typical vehicle reverse parking scenario

The flowchart of this modified Hybrid A* algorithm is illustrated in Figure 3. The algorithm works iteratively, generating multiple kinematically feasible path segments, referred to as motion primitives, at each iteration by simulating vehicle kinematics over a fixed time horizon. From the set of partial paths already formed in earlier iterations, new motion primitives are initiated from the node at the end of the path that is currently closest to the goal. Closeness to the goal is evaluated using a cost metric, where a lower cost indicates an increased proximity to the target. After these motion primitive branches are created, they are checked for collisions. The endpoints of those that remain collision-free become the new terminal nodes of their respective partial paths. Among these, the partial path with the lowest cost is selected to be expanded in the next iteration. This process repeats until a branch approaches the goal sufficiently, at which point that path is accepted as the final solution that is both kinematically valid and collision-free.

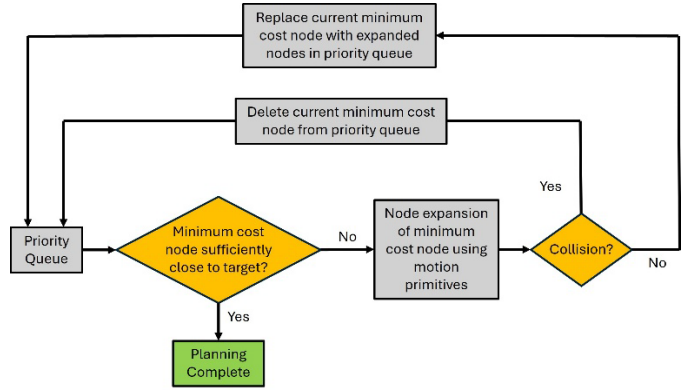

Figure 3. Modified Hybrid A* algorithm flowchart

Information about each partial path, such as the system states at its terminal nodes and their associated cost values, is maintained in a priority queue. This queue is ordered by cost, ensuring that the node with the lowest cost can be quickly identified for subsequent motion primitive branch expansion. Once the newly generated admissible nodes have been inserted into the queue after each iteration, the entry from which they were spawned is removed, as its information has already been incorporated into the new nodes. Additionally, if the node with the lowest cost in the queue fails to produce any collision-free branches, the algorithm will proceed to evaluate the next best candidate, i.e., the node with the second-lowest cost, for branch expansion.

## Cost Function

As noted in the algorithm overview, cost values are assigned to the end nodes of partial paths to assess their proximity to the desired final state. This sub-section aims to provide further insight into the design of the cost function.

In this path-planning framework, the cost function consists of two primary components: a heuristic term and an accumulated action cost term. The heuristic term estimates how closely the vehicle states at a given end node align with the target states. It is formulated in a quadratic form as

$$J_H = \left( \begin{bmatrix} X_R \\ Y_R \\ \psi \end{bmatrix} - \begin{bmatrix} X_{R,goal} \\ Y_{R,goal} \\ \psi_{goal} \end{bmatrix} \right)^T Q \left( \begin{bmatrix} X_R \\ Y_R \\ \psi \end{bmatrix} - \begin{bmatrix} X_{R,goal} \\ Y_{R,goal} \\ \psi_{goal} \end{bmatrix} \right) \quad (10)$$

where the $Q$ matrix is a $3 \times 3$ positive definite matrix.

Beyond the heuristic term, additional factors must be considered when selecting which partial path to expand. The necessity of this step is illustrated in Figure 4, where Path 1 reaches a given query node directly, while Path 2 passes through the same node but completes a full loop before returning to it. Although both paths terminate at identical states and, therefore, share the same heuristic cost, Path 1 is inherently more efficient because it reached the state using fewer expansion steps. To account for this distinction, an additional cost component, referred to as the accumulated action cost $J_A$, is introduced as

$$J_A = K_A \cdot N_A \quad (11)$$

where $N_A$ denotes the number of actions already taken, and $K_A$ is a scaling parameter. By combining this accumulated action cost with the heuristic term, the full cost function is expressed in Equation (12).

$$J = J_H + J_A \quad (12)$$

It is important to emphasize that the scaling parameter $K_A$ must be chosen such that $J_A$ remains significantly smaller than the heuristic cost $J_H$, as its primary role is to distinguish between partial paths with identical end states without distorting the evaluation of goal proximity of the heuristic cost term.

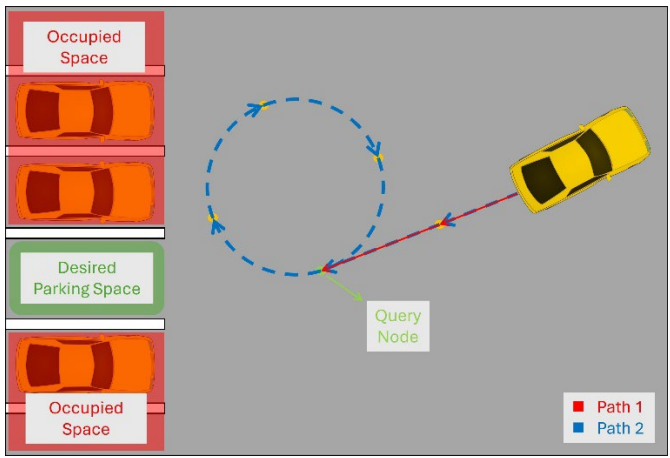

Figure 4. Example of identical heuristic cost

## Motion Primitives

As discussed in the algorithm overview, each iteration involves generating kinematically feasible motion primitives, where multiple branches are created from a single node. Since this work focuses on a vehicle system, these branches are produced by simulating the kinematic single-track vehicle model over a fixed time interval. Although the number of branches can be custom-defined, this study adopts a five-branch configuration. These five branches represent the following maneuvers: 1) reversing with maximum left steering; 2) reversing with moderate left steering; 3) reversing with zero steering; 4) reversing with moderate right steering; 5) reversing with maximum right steering. All branches use the same vehicle rear axle reverse speed $V_R$ (negative in value), and with an identical simulation duration, their terminal nodes can be directly compared using the cost function defined earlier. Figure 5 provides an illustration of these motion primitive branches. It should also be noted that each branch maintains a constant steering angle during the branch expansion process.

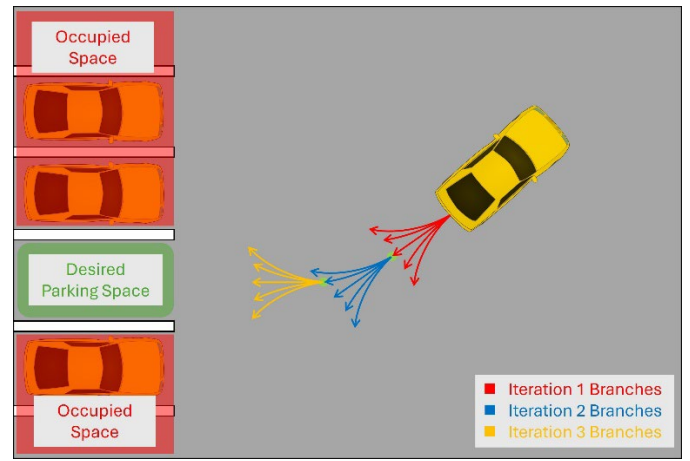

Figure 5. Motion primitive branches

## Collision Check

After generating the motion primitives, each candidate branch undergoes a collision check to verify whether the kinematically feasible segments can also avoid obstacles within the parking environment. To perform this assessment, a binary occupancy map is first constructed, defining both obstacle-occupied areas and the free regions available for vehicle motion. It is important to note that the motion primitives produced during the node expansion process represent only the trajectories of vehicle rear axle center. As a result, even if this trajectory clears the obstacles, parts of the vehicle body may still collide. To address this, the obstacle regions in the occupancy map are inflated by the half width of the vehicle, and these inflated regions are also designated as occupied. Figure 6 illustrates an example of such an inflated occupancy map. With this configuration, as long as the vehicle centerline remains within the green unoccupied region, collision avoidance is assured.

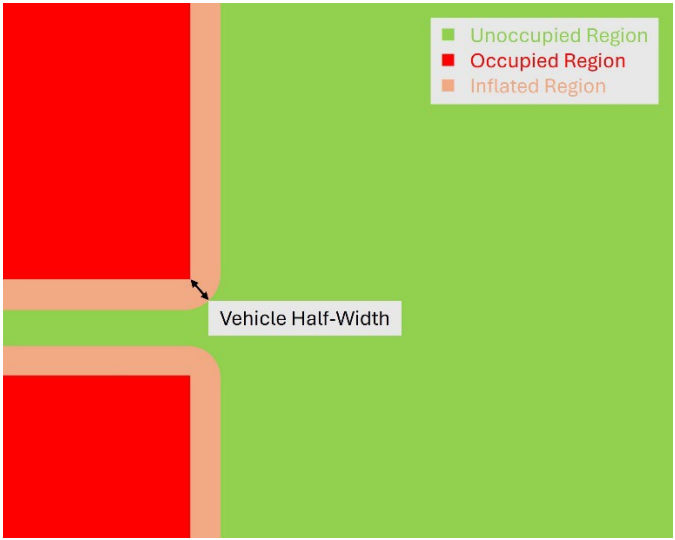

Figure 6. Example of an inflated binary occupancy map

With the binary occupancy map in place, the next task is to reconstruct the vehicle centerline. As the motion primitives already contain vehicle states $(X_R, Y_R, \psi)$, additional centerline points can be computed using the transport formula, the form of which is described in Equations 1 and 2. An illustration of this reconstruction process is shown in Figure 7. The resulting centerline is then evaluated against the inflated binary occupancy map to detect potential collisions. It is important to ensure that the reconstructed points are sufficiently dense, otherwise the vehicle centerline could intersect inflated obstacle regions without any sampled points falling within them, leading to undetected collisions.

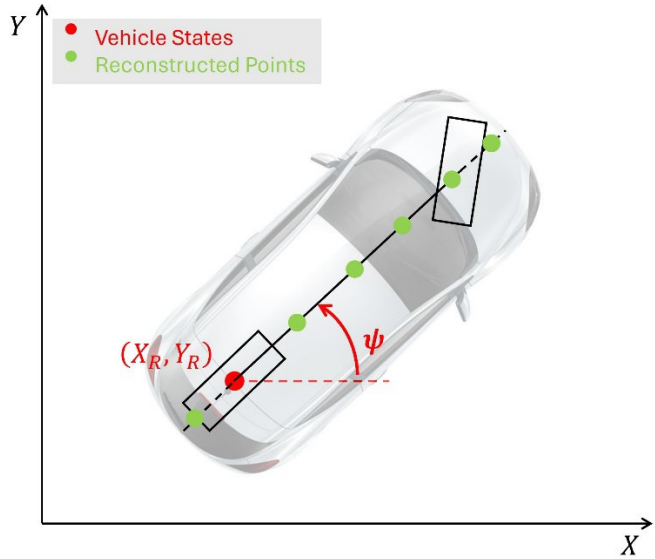

Figure 7. Vehicle centerline reconstruction

### Priority Queue

As discussed earlier, the priority queue stores information for all partial paths. In this proposed planning method, each queue entry contains the data summarized in Table 2. Among these elements, the action sequence, cost value, and terminal node states are essential for the planning process. The action sequence records the number of actions already taken and is used in the cost calculation. The cost itself determines which node should be expanded first, and the terminal node states serve as the initial conditions for further expansion. The remaining fields are included primarily to facilitate visualization and interpretation of the path-planning procedure. It is important to note that although the input history generated during the planning process could, in principle, be replayed to reproduce the planned trajectory, the overall system operates in an open-loop manner. Consequently, it is not inherently robust to external disturbances. A feedback controller would, hence, be necessary for accurate path tracking, but such control design lies beyond the scope of this paper.

Table 2. Priority queue components

| Queue Content | Explanation |
| --- | --- |
| Action Sequence | Sequence of actions already applied in the current partial path |
| Cost | Cost value of the terminal node in the current partial path |
| Terminal Node States | Vehicle states at the terminal node of current partial path |
| Expanded Branch Trajectory | Vehicle trajectory of the most recent motion primitive branch in the current partial path |
| Overall Path Trajectory | Overall vehicle trajectory of the current partial path |
| Input History | Vehicle inputs history for the current partial path |

## Simulation Study

Simulation studies are conducted to demonstrate the effectiveness of the modified Hybrid A* planning algorithm described in this paper. The vehicle geometric parameters and values of additional planning settings are presented in Table 3. It should be remarked that the five motion primitive branches generated at each algorithm iteration take different steering input values from the $\delta_f$ value choices However, they all share the same simulation duration and vehicle rear axle speed $V_R$ so that they are comparable to each other.

Table 3. Parameter value choices for simulation study

| Model Parameter | Value Choice |
| --- | --- |
| $L$ | 2.896 [m] |
| Vehicle Length | 4.878 [m] |
| Vehicle Width | 1.935 [m] |
| Node Expansion Simulation Duration | 1 [sec] |
| $Q$ | $\begin{bmatrix} 1 & 0 & 0 \\ 0 & 5 & 0 \\ 0 & 0 & 1 \end{bmatrix}$ |
| $K_A$ | 0.1 |
| $V_R$ | -1 [m/sec] |
| $\delta_f$ Choices | [-0.75, -0.35, 0, 0.35, 0.75] [rad] |

The binary occupancy map used for the simulation case study is shown in Figure 8. The map is constructed to replicate a mostly occupied parking lot with limited spaces for maneuvering. The goal of the path-planning operation is hence to position the vehicle into the one remaining parking space while avoiding any collisions with other vehicles parked in the neighboring spaces as well as across the driveway. All occupied regions in the occupancy map are also inflated with vehicle half-width in order to facilitate collision avoidance functions described earlier in this paper.

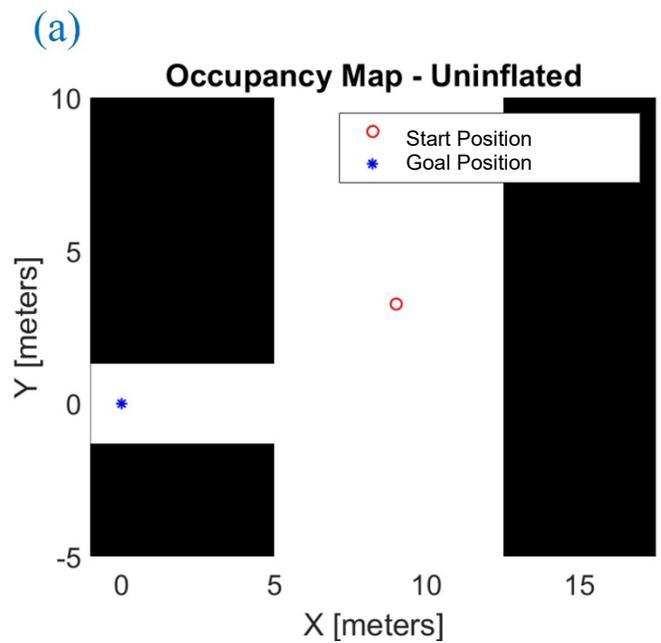

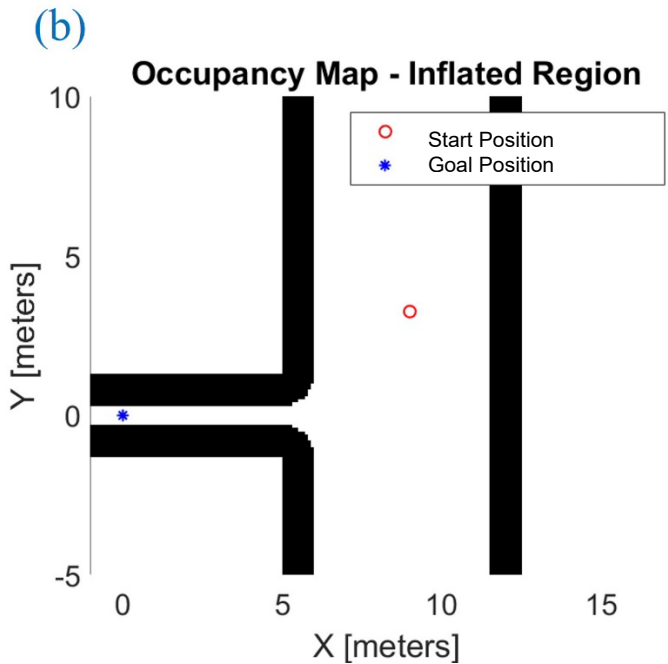

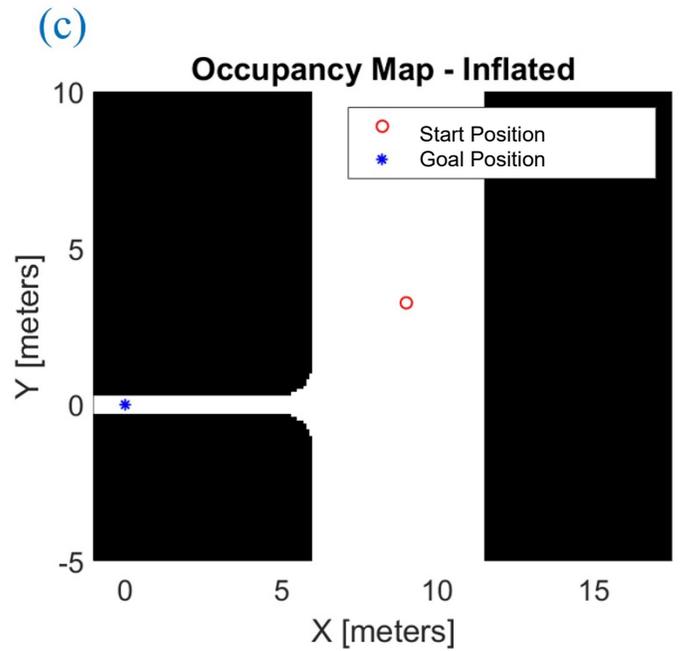

Figure 8. Binary occupancy map construction: (a) Original uninflated map; (b) Inflated map regions; (c) Modified inflated map

The results of the case study are shown in Figure 9 to Figure 11. Figure 9 records all kinematically feasible and collision-free motion primitive branches generated and stored in the priority queue during the entire planning process. It can be observed that among all the motion primitive branches in the priority queue, one sequence of branches can be joined together to construct a complete path that leads from the start point to the end point, effectively achieving the planning goal. Figure 10(b) illustrates this planned path more clearly, and Figure 10(a) displays the motion of the entire vehicle following this trajectory. Additionally, Figure 11 shows the vehicle front axle steering angle required to replicate the planned path, and it can be observed that the steering input does not exceed the lower and upper limits of $\delta_f$. Many more (spatially feasible) vehicle initial conditions have been tested with the proposed path-planning algorithm, and all of them demonstrated satisfactory performance in generating collision-free paths that guide the vehicle from the initial position and orientation into the designated parking space.

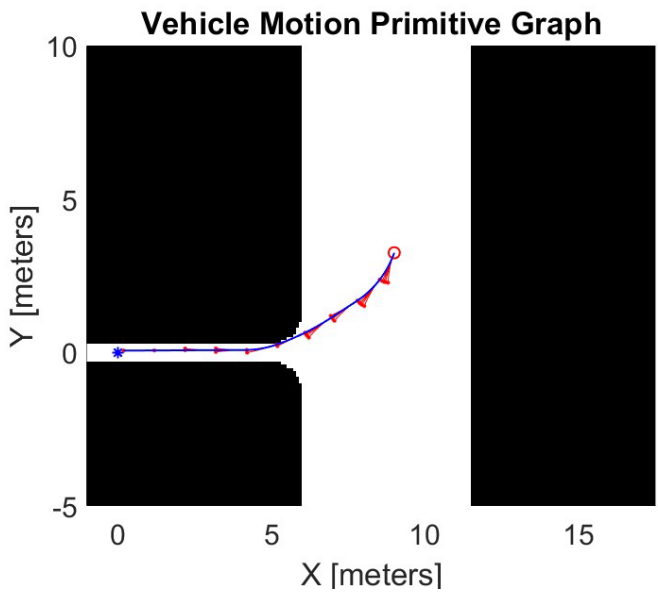

Figure 9. Path-planning simulation result: motion primitive graph

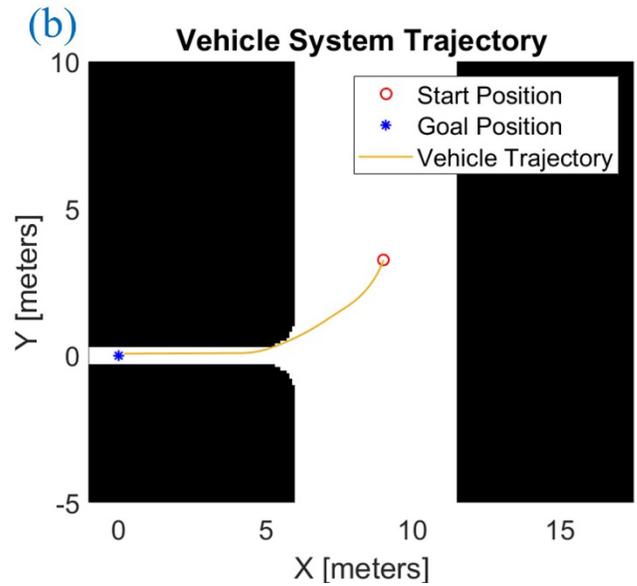

Figure 10. Path-planning simulation results: (a) Vehicle reverse motion; (b) Vehicle rear axle center trajectory

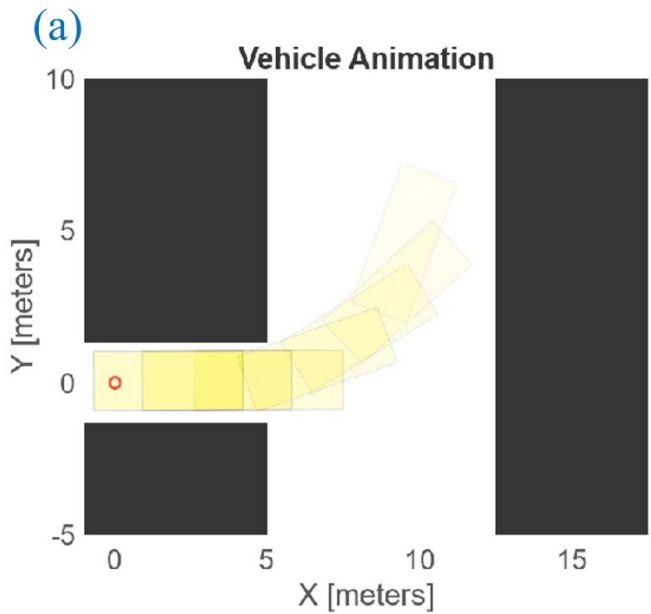

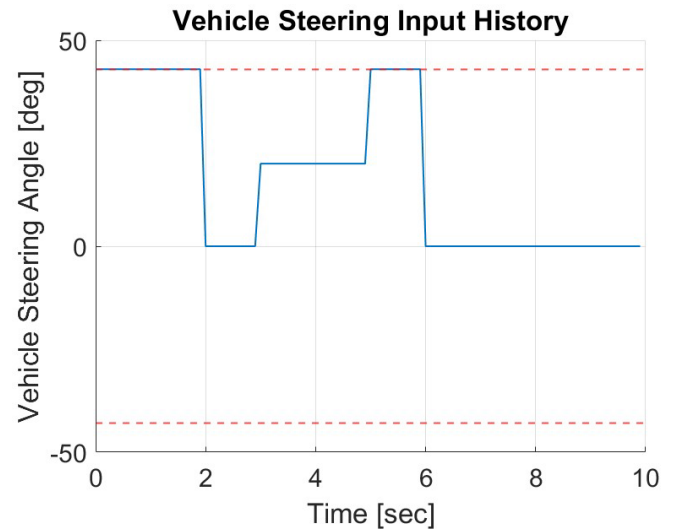

Figure 11. Path-planning simulation result: vehicle front axle steering angle

## Conclusions

This paper presented a modified Hybrid A* path-planning algorithm to handle vehicle reverse parking maneuvers in tight spaces. A kinematic single-track vehicle model is first derived to be used as the motion model in the proposed planning algorithm. The modified Hybrid A* path-planning formulation is designed such that kinematic feasibility and static obstacle avoidance capability can be guaranteed simultaneously. A simulation study was used to demonstrate the effectiveness of the proposed planning approach. Future work will include the integration of dynamic models into the formulation to enhance fidelity as well as the addition of a closed-loop path-tracking controller to follow the planned path. Future work will include research in decision making based on deep reinforcement learning [36], delay tolerance [37], and the use of control barrier functions to add more safety boundaries [38], [39] during operation.